\documentclass[10pt,twocolumn,letterpaper]{article}

\usepackage[pagenumbers]{cvpr} % To force page numbers, e.g. for an arXiv version

\usepackage{booktabs}
\usepackage{fontawesome5}
\usepackage{roboto-mono}
\usepackage{makecell}
\usepackage{microtype}
\usepackage{soul}
\usepackage{tabularx}
\usepackage{xspace}

\providecommand{\ci}[1]{{\color{gray}\tiny{$\,\pm$#1}}}

\usepackage{amsmath,amsfonts,bm}

\def\eqref#1{equation~\ref{#1}}
\def\1{\bm{1}}

\DeclareMathAlphabet{\mathsfit}{\encodingdefault}{\sfdefault}{m}{sl}
\SetMathAlphabet{\mathsfit}{bold}{\encodingdefault}{\sfdefault}{bx}{n}

\newcommand{\model}{\textsc{OmniAct}\xspace}
\newcommand{\modelshort}{\textsc{OmniAct}\xspace}
\newcommand{\modelbench}{\textsc{OmniAct-Bench}\xspace}

\definecolor{cvprblue}{rgb}{0.21,0.49,0.74}
\usepackage[pagebackref,breaklinks,colorlinks,allcolors=cvprblue]{hyperref}

\def\confName{CVPR}
\def\confYear{2027}

\title{OmniAct: One Policy, One Million Manipulators}

\author{%
  Xiang Fan$^{1*}$\quad
  Yuheng Wang$^{1*}$\quad
  Jiayuan Yan$^1$\quad
  Bohan Fang$^1$\quad
  Wisdom Oluchi Ikezogwo$^1$ \\
  Ainaz Eftekhar$^1$\quad
  Max Argus$^3$\quad
  Zhongzheng Ren$^{1,2}$\quad
  Jiafei Duan$^4$\quad
  Ranjay Krishna$^1$ \vspace{0.2em}\\
  $^1$University of Washington\quad
  $^2$University of North Carolina at Chapel Hill\quad\\
  $^3$Allen Institute for Artificial Intelligence\quad
  $^4$National University of Singapore \vspace{0.2em}\\
  $^*$Equal contribution \vspace{0.2em} \\
  {\color[HTML]{E8A317}\faEnvelope}\ \texttt{xiangfan@cs.washington.edu} \\
  {\color[HTML]{367DD4}\faGlobe}\ \url{https://omniact.github.io/}
}

\begin{document}

\twocolumn[{
\maketitle
\begin{center}
  \includegraphics[width=\linewidth]{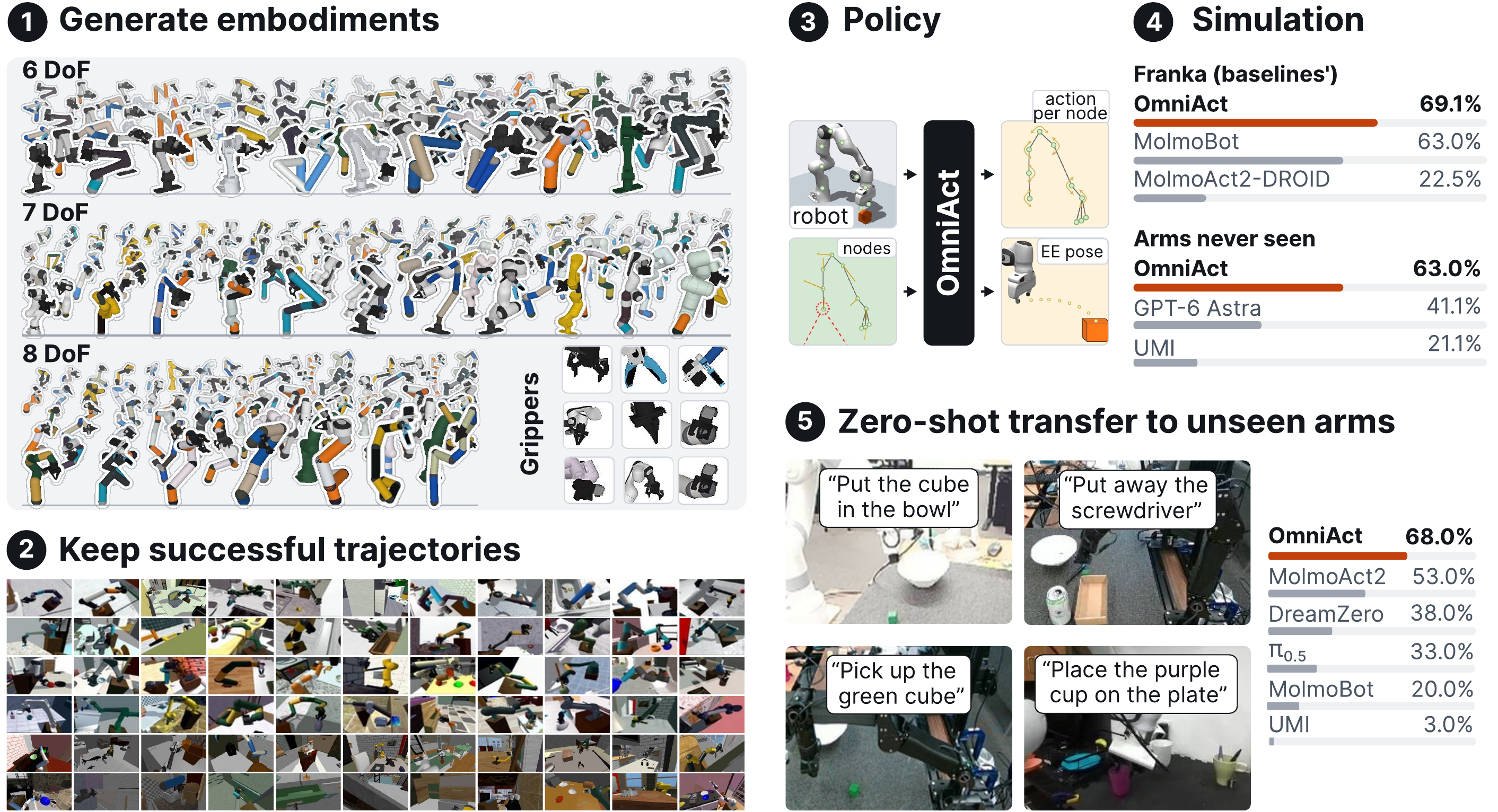}
  \captionof{figure}{\textbf{\model is an embodiment-native policy} designed to employ robot-specific motion strategies without embodiment-agnostic planning. We generate (1)~diverse embodiments procedurally and (2) varied simulated trajectories at scale and (3)~train a single policy, \model. (4/5)~In simulation and real, \model outperforms all baselines (zero shot) on seen and unseen embodiments.}
  \label{fig:teaser}
\end{center}
}]

\begin{abstract}
A single policy that controls diverse robots must not only transfer task knowledge across embodiments, but also adapt its behavior to each robot's capabilities. Existing approaches facilitate transfer through shared action representations, but these representations do not necessarily capture how differences in robot morphology should shape manipulation strategies.
We introduce \model, a \emph{cross-embodiment} manipulation policy that conditions directly on each robot's \textit{kinematics in a specific scene} to produce embodiment-specific behavior. 
This kinematic graph represents the robot's joint positions, motion axes, and connectivity to its visual observation, allowing the policy to account for how its embodiment relates to objects and surfaces in the scene.
A flow-matching policy predicts relative joint motions for each node, enabling control across robots with different degrees of freedom and kinematic configurations.
To learn how physical structure shapes behavior at scale, we procedurally design and filter one million physically feasible robot embodiments; next, we develop algorithms to generate robot trajectories in simulation for all embodiments; finally, we train across diverse simulated manipulation tasks.
On 2,000 embodiment-scene-held-out evaluation episodes, \model achieves 63.0\% success, compared to 17.8\% for $\pi_{0.5}$ and 57.9\% for the strongest baseline; on the task families GPT-6 Astra was evaluated on, it achieves 64.7\% against 41.1\%. On a previously unseen real-world robot, it achieves nearly twice the success rate of the next best method, MolmoAct2. Beyond task success, we observe that \model learns distinct manipulation strategies that exploit the capabilities and accommodate the limitations of different morphologies.
\end{abstract}

\section{Introduction}
\label{sec:intro}

% STATE OF THE WORLD and PROBLEM WITH THE STATE
A single policy capable of controlling robots with diverse embodiments must solve two related problems: \textit{transferring task knowledge across robots} and \textit{adapting behavior to each robot's physical capabilities}. Existing approaches have made substantial progress on the former by identifying representations and interfaces that can be shared across embodiments. Generalist vision-language-action models use shared end-effector commands, unified action tokenizers, or learned action latents~\cite{embodimentcollaboration2025openxembodimentroboticlearning,octomodelteam2024octoopensourcegeneralistrobot,kim2024openvlaopensourcevisionlanguageactionmodel,liu2025rdt1bdiffusionfoundationmodel,black2026pi0visionlanguageactionflowmodel}. Object-centric methods represent desired object motions or interactions that can be transferred between robots~\cite{bharadhwaj2024track2actpredictingpointtracks,yuan2024generalflowfoundationaffordance,huang2024rekepspatiotemporalreasoningrelational,liu2024mokaopenworldroboticmanipulation}. These approaches facilitate transfer by exploiting commonalities across embodiments. However, transferring task knowledge does not necessarily lead to effective control of each robot. Two robots pursuing the same goal may require different grasps, trajectories, or contact strategies because of differences in their kinematics, workspace, and dexterity.

% WHY IS IT A CHALLENGING PROBLEM?
Learning embodiment-specific behaviors presents several challenges. 
\textit{First}, a policy must reason about the robot's geometry in relation to the observed scene. The usefulness of a joint configuration depends on where its joints and motion axes lie relative to the objects and obstacles it must interact with.
Existing policies incorporate descriptions such as URDFs or kinematic graphs, but often encode the robot independently of the scene~\cite{chen2019hardwareconditionedpoliciesmultirobot,patel2024getzerographembodimenttransformer,suzuki2026embeddingmorphologytransformerscrossrobot}. 
\textit{Second}, the policy must accommodate different kinematic structures, including variations in joint types, degrees of freedom, and link topology, while still supporting a common control interface. Existing approaches have demonstrated transfer across relatively small collections of manipulators~\cite{zhao2026geopropgroundingrobotstate,vosylius2024renderdiffusealigningimage,lv2025kinematicsawarediffusionpolicyconsistent}. 
\textit{Third}, the policy must learn how changes in morphology affect behavior so that it can control embodiments not encountered during training. This requires experience across a sufficiently diverse set of physical structures. Prior efforts to scale learning across hundreds of morphologies have primarily focused on locomotion~\cite{ai2025embodimentscalinglawsrobot,gupta2022metamorphlearninguniversalcontrollers} and not manipulation.

% INSIGHT AND TECHNICAL CONTRIBUTION
We introduce \textbf{\model}, a policy trained across one million embodiments, capable of adapting behavior to each embodiment and what it can achieve in a given scene. 
Our approach represents each robot through a \textit{scene-grounded kinematic graph}, expressing its joint geometry in the 3D coordinate frame of each observing camera. Each joint and fingertip is represented as a node containing its camera-frame position, motion axis, joint type, proprioceptive state, and kinematic connectivity. This allows the policy to relate the robot's physical structure to the corresponding visual observations.
We pair this representation with a \textit{variable-DoF action representation} that jointly predicts relative joint motions and end-effector pose updates using a flow-matching policy. By unifying the indexing scheme across all embodiments, our model learn to support different numbers and configurations of joints within the same network.
These two components allow the policy to generate motions directly from each robot's kinematic structure and its relationship to the scene.

% DATA CONTRIBUTION.
Training a policy to exploit these differences requires substantial diversity in robot morphology. A policy trained on a handful of embodiments may learn effective manipulation strategies for those robots without learning how to adapt them to new ones. 
We develop a procedure to \textbf{instantiate one million unique robot embodiments and create algorithms to automatically generate trajectories} for each of them in simulation.
Our data generation pipeline varies joint configurations, link geometry, and kinematic topology to generate embodiments; we use inverse-kinematics screening and physics-based acceptance to ensure feasibility of each trajectory. The resulting training distribution exposes the policy to a broad range of robot configurations and allow us to test on unseen embodiments.

% EXPERIMENTS
We evaluate \model on embodiments and scenes held out from training. On \modelbench, it succeeds in 63.0\% of episodes within 500 steps, against 57.9\% for the strongest baseline and 17.8\% for $\pi_{0.5}$; on the task families GPT-6 Astra was evaluated on, it reaches 64.7\% against GPT-6 Astra's 41.1\% at the same step budget and 54.8\% with twice the budget (\Cref{tab:molmospaces_benchmark_results}). On a real OpenYAM arm with a fine-tip
gripper, neither seen in training, \model completes 13 of 20 pick-and-place trials. The only baseline to succeed at all is MolmoAct2-BimanualYAM, for which this arm is in distribution, at 7 of 20; every other baseline completes none (\Cref{tab:real_world_results}). We observe that on this arm the policy sometimes makes use
of the long final link specific to this unseen embodiment, extending it horizontally and then rotating the gripper to reach
far targets (\Cref{sec:motion_strategy}), a behavior not seen from MolmoAct2 trained exclusively on YAM. These results suggest that policies grounded in a robot's kinematics are a path towards better cross-embodiment generalization.

\nocite{octomodelteam2024octoopensourcegeneralistrobot,wang2024scalingproprioceptivevisuallearningheterogeneous,black2026pi0visionlanguageactionflowmodel,bu2025univlalearningacttaskcentric,chi2024universalmanipulationinterfaceinthewild,seo2025legatocrossembodimentimitationusing,ma2024hierarchicaldiffusionpolicykinematicsaware,nvidia2025gr00tn1openfoundation,doshi2024scalingcrossembodiedlearningpolicy,wen2025dexvlavisionlanguagemodelplugin,gupta2022metamorphlearninguniversalcontrollers,patel2024getzerographembodimenttransformer,suzuki2026embeddingmorphologytransformerscrossrobot,ha2026transformertransformerunifiedmodel,zhang2026dexformercrossembodieddexterousmanipulation,xiaomiembodiedintelligenceteam2026policyembodimentsunifiedcameracentric,zhao2026geopropgroundingrobotstate,vosylius2024renderdiffusealigningimage,lv2025kinematicsawarediffusionpolicyconsistent,chen2024miragecrossembodimentzeroshotpolicy,niu2026skelwamskeletonguidedworldactionmodel,lepert2025masqueradelearninginthewildhuman,kim2026robotfactoredworldmodelsrobot,goswami2025robopeppvisionbasedrobotpose,liu2024differentiablerobotrendering,embodimentcollaboration2025openxembodimentroboticlearning,contributors2026agibotworldcolosseolargescale,rath2025xmopwholebodycontrolpolicy}

\section{Method}
\label{sec:method}

We introduce \model, a single policy that learns cross-embodiment manipulation across robots with different kinematic structures. Our approach has three components. First, we represent each robot through a scene-grounded kinematic graph that expresses its geometry in the coordinate frame of its visual observations. Second, we develop a variable-DoF action representation that jointly predicts joint motions and end-effector poses across different robot configurations. Third, we generate diverse embodiments and collect successful manipulation trajectories in simulation to train the policy across one million robots.

\begin{figure}[t]
  \centering
  \includegraphics[width=\linewidth]{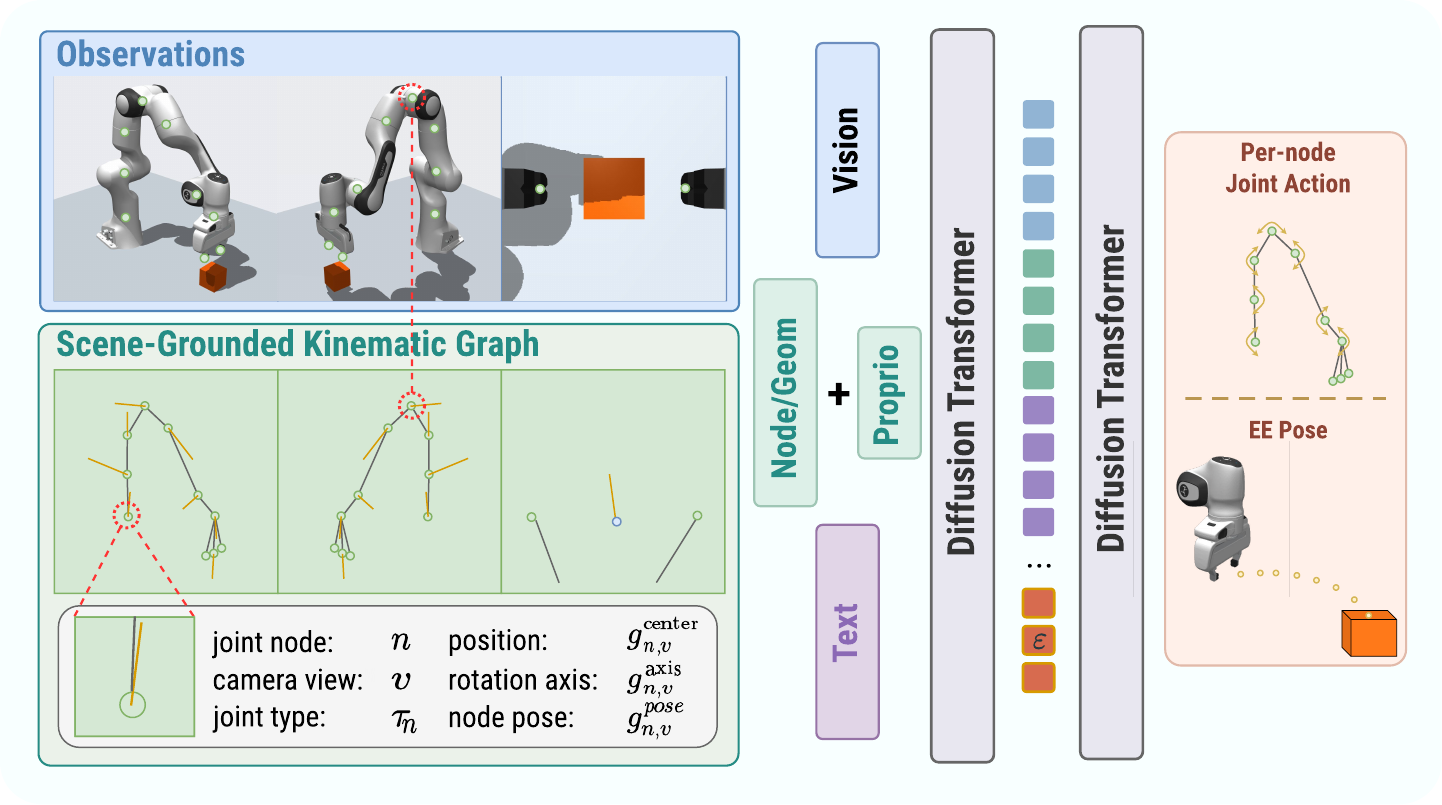}
  \caption{\textbf{\model architecture.} Multi-view RGB observations, language instructions, and camera-frame scene-grounded kinematic graph condition a shared flow-matching Transformer. Each node represents a joint or annotated point on the robot, encoding its geometric and proprioceptive state in each camera frame. The policy jointly predicts relative joint motions and end-effector poses through a variable-DoF action representation.}
  \label{fig:arch}
\end{figure}

\paragraph{Policy formulation.}
We consider manipulation tasks across robots with different numbers of joints, link geometries, and kinematic structures. At each time step $t$, the policy receives a language instruction $\ell$, multi-view RGB observations $\mathcal{I}_t=\{I_{t,v}\}_{v=1}^{V}$, and the robot's current configuration and proprioceptive state. The robot's embodiment specifies its kinematic topology, joint types, limits, and geometric parameters.

The policy predicts an action chunk containing relative joint motions and end-effector (EE) pose updates. Both predictions are supervised during training and used for control during execution. A prefix of the predicted chunk is executed before obtaining updated observations and replanning. The same policy parameters are shared across all embodiments.

We encode RGB observations using SigLIP v2~\citep{tschannen2025siglip2multilingualvisionlanguage} and language instructions using Qwen 2.5~\citep{bai2025qwen25vltechnicalreport}. Robot geometry, configuration, and proprioception are represented through kinematic node tokens.

\subsection{Scene-Grounded Kinematic Graph}
\label{sec:method_embodiment}

The motions available to a robot depend on its kinematic structure and how that structure is situated relative to the visual observation.
% We represent each robot as a kinematic graph whose nodes encode its joints, fingertips, and other relevant points. Each node carries information about its kinematic relationships, physical configuration, and current geometric state. 
% We express this str in each observing camera's 3D coordinate frame, allowing the policy to relate the robot's joints and motion axes to the corresponding visual observations.

\paragraph{Kinematic structure.}
We decompose each robot into nodes representing arm joints, gripper drivers, gripper tips, and annotated points such as the tool center point (TCP). Each node $n$ has a type
$\tau_n\in\{\mathrm{terminal},\mathrm{revolute},\mathrm{prismatic},\mathrm{point}\}$
and an actuation indicator that distinguishes driven joints, terminal joints, and annotated points. The kinematic topology is encoded through shortest-path distances between nodes, which enter the Transformer as attention biases~\citep{ying2021transformersreallyperformbad}.
Nodes are ordered according to their kinematic structure and left-padded to a maximum supported node count, keeping end-effector nodes at the final positions of the sequence. This provides a consistent indexing scheme across different embodiments while preserving their variable numbers of joints.

\paragraph{Joint proprioception.}
Joint states differ in their physical units, limits, and ranges across embodiments. We center each joint coordinate around the midpoint of its limits and convert it to a common numerical scale:
$p_n = \kappa_n(q_n-c_n)$,
where $q_n$ is the current joint position, $c_n$ is the midpoint of its limit range, and $\kappa_n$ is a joint-type-dependent unit factor. Revolute joints are expressed in radians and prismatic joints in decimeters. The centered scalar is expanded into Fourier features and encoded through an MLP.

\paragraph{Camera-frame geometric grounding.}
For each node $n$ and camera $v$, we compute its current geometry using forward kinematics and known camera extrinsics. The geometry is represented in metric 3D camera coordinates through three fields:
$G_{n,v} =
    \left(
    g^{\mathrm{center}}_{n,v},
    g^{\mathrm{axis}}_{n,v},
    g^{\mathrm{pose}}_{n,v}
    \right)$,
where $g^{\mathrm{center}}_{n,v}\in\mathbb{R}^3$ denotes the node center, $g^{\mathrm{axis}}_{n,v}\in\mathbb{R}^3$ is a point one unit along its motion axis, and $g^{\mathrm{pose}}_{n,v}\in\mathbb{R}^9$ represents translation and orientation using a continuous 6D rotation representation~\cite{zhou2020continuityrotationrepresentationsneural}. The motion axis describes the rotation axis of a revolute joint or the translation direction of a prismatic joint.

Centers are provided for all nodes, motion axes for actuated nodes, and poses for annotated nodes. Unavailable fields are zeroed before encoding. In single-arm manipulation, the TCP is represented as an annotated descendant of the wrist, allowing its pose to be incorporated into the same representation.

The camera-frame coordinates are encoded through sinusoidal features and a learned projection, then combined with proprioception and discrete node attributes:
\begin{multline}
    z^G_{n,v}
    =
    f_G\left(
        g^{\mathrm{center}}_{n,v},
        g^{\mathrm{axis}}_{n,v},
        g^{\mathrm{pose}}_{n,v}
    \right)
    + f_P(p_n)
    \\
    + e^{\mathrm{slot}}_{j(n)}
    + e^{\mathrm{type}}_{\tau_n}
    + e^{\mathrm{view}}_v
    + e^{\mathrm{node}},
    \label{eq:grounded_node_token}
\end{multline}
where $f_G$ and $f_P$ encode geometry and proprioception, $e^{\mathrm{slot}}_{j(n)}$ identifies the node's padded slot, and $e^{\mathrm{type}}_{\tau_n}$ identifies its type. The embedding $e^{\mathrm{node}}$ distinguishes node tokens from other modalities.
We construct one token per node and camera view. Each token shares its view embedding $e^{\mathrm{view}}_v$ with image tokens from the corresponding camera, allowing attention to associate the robot's geometry with its visual observations and integrate information across views.

Geometric grounding is performed in the camera's 3D coordinate frame. We do not explicitly project joint positions into image pixels or sample visual features at projected locations. The shared viewpoint and view embeddings provide correspondence between geometric and visual tokens, while the Transformer learns their interactions.

\subsection{Variable-DoF Action Prediction}
\label{sec:method_action_rep}

Robots differ in their numbers of joints, joint types, and available motion directions. We use a node-aligned action representation in which the output follows the same padded ordering as the input kinematic graph. This allows a single network to predict actions for different embodiments while maintaining correspondence between each node's geometry and its control signals.

Each output node is processed by two prediction heads: a relative joint-motion head and a relative pose head. The joint-motion head predicts changes in the robot's joint coordinates, while the pose head predicts relative EE poses associated with the corresponding end-effector nodes. Both heads are trained through flow matching, and both predictions contribute to the control executed by the robot.

We normalize joint actions using empirical statistics associated with their left-padded node slots, which we find effective for representing diverse embodiments. Action outputs are masked according to the valid nodes and action quantities for each robot.

\paragraph{Architecture and training.}
\label{sec:method_arch}

Visual tokens, language tokens, grounded kinematic nodes, and noisy action tokens are jointly processed by a Transformer. The model is trained using flow matching~\citep{lipman2023flowmatchinggenerativemodeling}. Let $y$ denote a normalized target action chunk containing joint motions and EE pose updates. At flow time $s\in[0,1]$, we construct
\begin{equation}
    x_s=(1-s)\epsilon+sy,
    \qquad
    \epsilon\sim\mathcal{N}(0,I),
    \label{eq:flow_interpolation}
\end{equation}
where Gaussian noise is sampled over valid action dimensions.

The network predicts the velocity of this interpolation, conditioned on the language instruction, visual observations, and scene-grounded kinematic graph. We train using the flow-matching objective over valid action dimensions:
\begin{equation}
    \mathcal{L}_{\mathrm{FM}}
    =
    \mathbb{E}_{s,\epsilon,y}
    \left[
    \left\|
    M\odot
    \left(
    v_\theta(x_s,s;\ell,\mathcal{I}_t,G_t)
    -(y-\epsilon)
    \right)
    \right\|_2^2
    \right],
    \label{eq:flow_loss}
\end{equation}
where $M$ masks invalid or padded action dimensions and $G_t$ denotes the grounded kinematic representation.
The model learns joint and EE pose predictions jointly. At inference, the predicted action chunk provides both joint-level motions and EE pose updates to the control interface. A prefix of the chunk is executed before the policy replans from updated observations.

\subsection{Scaling Embodiments}
\label{sec:method_scaling_embodiments}

\paragraph{Embodiment generation.} A policy trained on a small set of robots may learn manipulation strategies specific to those kinematics, limiting its ability to adapt to unfamiliar embodiments. We therefore procedurally generate one million robot embodiments spanning different kinematic structures, joint configurations, and link geometries.
Each embodiment is generated deterministically from a random seed, generating its morphology, joint limits, actuator dynamics, link geometry, inertia, and appearance. This makes each robot reproducible without separately storing its complete description.

We filter generated candidates for self-collision and grasp feasibility. Grasp feasibility is evaluated by solving inverse kinematics for representative target grasp poses on randomly sampled objects. 84.4\% of all generated embodiments fail self-collision and 97.5\% of the surviving embodiments fail grasp checks despite being sampled from distributions designed to produce realistic robots. 6-DoF, 7-DoF, and 8-DoF embodiments have pass rates of 0.016\%, 0.299\%, and 0.653\% overall pass rate. Filtering these candidates avoids expensive trajectory collection for embodiments that cannot perform simple manipulation tasks.

\paragraph{Generating trajectories per embodiment.}
\label{sec:method_trajgen}
We collect manipulation trajectories in MolmoSpaces~\citep{kim2026molmospaceslargescaleopenecosystem}, instantiating each feasible embodiment in randomly generated scenes with different tasks and initial configurations. We plan two scene-task instances per embodiment.

Trajectory generation uses a planner that proposes locally feasible motions under the current robot's kinematic constraints. At each step, the planner considers feasible immediate motions towards the next desired pose~\cite{kim2026molmospaceslargescaleopenecosystem} without explicitly planning an embodiment-specific long-horizon strategy. Different from MolmoSpaces' use of top-down poses calculated assuming the Franka robot, we randomly rotate the grasp poses and allow for any kinematically feasible pose on the robot to form the support. We filter for successful episodes, producing demonstrations of manipulation behaviors that are feasible for each robot. 6-, 7-, and 8-DoF embodiments have flat episode success rates of 35.9\%, 34.8\%, and 36.4\%, compared to embodiment screening's DoF-dependent pass rates. Successful episodes are stored with sufficient simulation state to reconstruct the environment at intermediate steps. \Cref{tab:training_data} summarizes the resulting corpus.

A fixed-seed RNG is used to reproducibly hold out 5\% of embodiments and, within each task, 5\% of scenes. A trajectory is used for training only if neither its embodiment nor its scene is reserved. An embodiment or scene is called \emph{seen} when it appears in at least one accepted training trajectory. After this procedure, 18.3\% of embodiments and 10.9\% of scenes end up \emph{unseen} from training, directly or indirectly. 14.5\% of all trajectories are comprised of both unseen embodiments and unseen scenes. The subset with both embodiment and scenes unseen is denoted \modelbench, used for benchmarking.

\section{Experiments}
\label{sec:experiments}

\begin{table}[t]
    \caption{\textbf{\model-1M dataset statistics} on a total of $1,029{,}605$ trajectories over $800{,}196$ embodiments.}
    \label{tab:training_data}
    \footnotesize
    \setlength{\tabcolsep}{4pt}
    \centering
    {\scriptsize
    \begin{tabular}{lrrrr}
            \toprule
            \textbf{Task} & \textbf{Episodes} & \textbf{Frames} & \textbf{Avg.\ length (s)} & \textbf{Total length (h)} \\
            \midrule
            Pick           & 295.0k & 26.3M & \hphantom{0}5.8 & \hphantom{0,}476 \\
            Open           & 227.7k & 38.8M & 11.2 & \hphantom{0,}707 \\
            PnP color      & 141.8k & 25.4M & 11.7 & \hphantom{0,}462 \\
            Close          & 140.7k & 20.2M & \hphantom{0}9.4 & \hphantom{0,}367 \\
            Pick-and-place & 127.7k & 21.1M & 10.8 & \hphantom{0,}385 \\
            PnP next-to    & \hphantom{0}96.6k & 18.9M & 12.8 & \hphantom{0,}343 \\
            \midrule
            \textbf{Total} & \textbf{1,029.6k} & \textbf{150.4M} & \hphantom{0}\textbf{9.6} & \textbf{2,740} \\
            \bottomrule
        \end{tabular}}

    \vspace{0.7em}

    \begin{minipage}[t]{0.58\linewidth}
        \centering
        \resizebox{\linewidth}{!}{%
        \begin{tabular}[t]{lrrr}
            \toprule
            \textbf{Embodiment parameter} & \textbf{p5} & \textbf{p50} & \textbf{p95} \\
            \midrule
            Reach (m)                        &  0.86 &  1.26 &  1.74 \\
            Segment length (m)               &  0.00 &  0.14 &  0.41 \\
            Arm mass (kg)                    &  3.9  & 10.2  & 18.3  \\
            Link mass (kg)                   &  0.06 &  0.55 &  4.72 \\
            Joint range (deg)                & 94    & 205   & 322   \\
            Joint stiffness                  & 37    & 1216  & 4145  \\
            Stiffness, shoulder:wrist        &  1.8  &  4.3  & 45    \\
            Joint damping ratio              &  1.1  &  5.1  & 14.6  \\
            Gripper mount rotation ($^\circ$)  &  1.0  & 10.1  & 159   \\
            Pedestal height (m)              &  0.57 &  0.70 &  0.84 \\
            \bottomrule
        \end{tabular}}
    \end{minipage}\hfill
    \begin{minipage}[t]{0.40\linewidth}
        \centering
        \resizebox{\linewidth}{!}{%
        \begin{tabular}[t]{lr}
            \toprule
            \textbf{Embodiment property} & \textbf{Share} \\
            \midrule
            Arms mixing $\ge\!2$ link sources & 70.5\% \\
            Arms with a forearm roll          & 65.1\% \\
            Prismatic gripper drivers         & 49.6\% \\
            Joints with off-center limits     & 49.4\% \\
            Arms with a universal joint       & 44.2\% \\
            Joints that move only one way     & 28.4\% \\
            Arms mixing two robots' links     & 24.0\% \\
            Arms with three parallel axes     & 10.0\% \\
            Joints past a full turn           & 1.6\% \\
            \bottomrule
        \end{tabular}}
    \end{minipage}
\end{table}

    \begin{table*}[t]
    % side-by-side: tab:franka_sim (left) and fig:ms_pick_cumulative (right)
    \begin{minipage}[t]{0.655\textwidth}\vspace{0pt}
        \caption{\textbf{Evaluation on MolmoSpaces with a real robot unseen to \model vs.\ other models trained for the robot}. Success rate are measured over a sample of $2{,}622$ held-out MolmoSpaces episodes with a $500$-step cap. (\textbf{PnPC} and \textbf{PnPN} indicate pick-and-place with a color constraint and a ``next to'' target, respectively, and are not present in MolmoSpaces' simple scenes eval.)
        }
        \vspace{0.2em}
        \label{tab:franka_sim}
        \footnotesize
        \setlength{\tabcolsep}{2pt}
        \centering
        \begin{tabular}{l c c c c c c c}
            \toprule
            \textbf{Model} & \textbf{Pick} & \textbf{PnP} & \textbf{PnPC} & \textbf{PnPN} & \textbf{Open} & \textbf{Close} & \textbf{All} \\
            & \scriptsize $n\!=\!647$ & \scriptsize $n\!=\!474$ & \scriptsize $n\!=\!394$ & \scriptsize $n\!=\!304$ & \scriptsize $n\!=\!551$ & \scriptsize $n\!=\!252$ & \scriptsize $n\!=\!2{,}622$ \\
            \midrule
            \multicolumn{8}{c}{\textit{Simple scenes}} \\
            \midrule
            $\pi_{0.5}$-DROID & 36.4\ci{1.5} & 13.6\ci{1.1} & -- & -- & 22.7\ci{1.3} & 65.1\ci{1.5} & 33.8\ci{0.7} \\
            MolmoAct2-DROID & 48.3\ci{1.6} & 23.3\ci{1.3} & -- & -- & 11.7\ci{1.0} & 71.3\ci{1.4} & 37.9\ci{0.8} \\
            \midrule
            \multicolumn{8}{c}{\textit{Full scenes}} \\
            \midrule
            $\pi_{0.5}$-DROID & 29.5\ci{1.8} & 17.2\ci{1.7} & 14.1\ci{1.8} & 16.0\ci{2.1} & 43.1\ci{2.1} & 73.4\ci{2.8} & 28.2\ci{0.9} \\
            DreamZero-DROID & \hphantom{ }5.9\ci{0.9} & \hphantom{ }2.1\ci{0.7} & \hphantom{ }2.5\ci{0.8} & \hphantom{ }3.0\ci{1.0} & 20.5\ci{1.7} & 66.3\ci{3.0} & 13.2\ci{0.7} \\
            MolmoAct2-DROID & 27.4\ci{1.8} & 18.6\ci{1.8} & 10.4\ci{1.5} & 12.8\ci{1.9} & 19.0\ci{1.7} & \textbf{76.3}\ci{2.7} & 22.5\ci{0.8} \\
            MolmoBot & \textbf{90.4}\ci{1.2} & 78.9\ci{1.9} & 78.7\ci{2.1} & 42.4\ci{2.8} & 18.3\ci{1.6} & 61.1\ci{3.1} & 63.0\ci{0.9} \\
            \midrule
            \textbf{\modelshort} & 89.2\ci{1.2} & \textbf{79.3}\ci{1.9} & \textbf{79.2}\ci{2.0} & \textbf{57.9}\ci{2.8} & \textbf{43.2}\ci{2.1} & 52.8\ci{3.1} & \textbf{69.1}\ci{0.9} \\
            \bottomrule
        \end{tabular}
        \end{minipage}\hfill
    \begin{minipage}[t]{0.33\textwidth}\vspace{0pt}
        \centering
        \includegraphics[width=\linewidth]{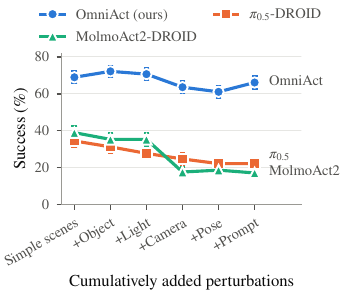}
        \captionof{figure}{\textbf{Cumulative robustness on MolmoSpaces Pick.} State-of-the-art policies $\pi_{0.5}$ and MolmoAct2 show decline in performance as more aspects of their input conditionings are diversified from the simple version.}
        \label{fig:ms_pick_cumulative}
    \end{minipage}
\end{table*}

\begin{table*}[t]
    \caption{\textbf{Evaluation on \modelbench with unseen robots}. \textbf{PnP}, \textbf{PnPN}, and \textbf{PnPC} are pick-and-place, pick-and-place next to a reference object, and pick-and-place by color. \textbf{Common} aggregates task families available on all evaluated models. \textbf{Steps} is the per-episode step cap. GPT-6 Astra is run on max thinking level. GPT-6 Astra replans every five steps under a budget of $200$ model calls, with a maximum allowed step count of $1{,}000$; the $200$- and $500$-step rows truncate those same episodes. Omitted GPT-6 tasks are due to limited API tokens.}
    \vspace{0.2em}
    \label{tab:molmospaces_benchmark_results}
    \footnotesize
    \setlength{\tabcolsep}{3pt}
    \centering
    \begin{tabular}{l c c c c c c c c c}
        \toprule
         \textbf{Model} & \textbf{Steps} & \textbf{Pick} & \textbf{PnP} & \textbf{PnPN} & \textbf{PnPC} & \textbf{Open} & \textbf{Close} & \textbf{Common} & \textbf{All} \\
        \midrule
        MolmoAct2-YAM & 500 & 0.2\ci{0.2} & 2.9\ci{1.0} & 0.0\ci{0.0} & 0.8\ci{0.6} & 3.0\ci{0.8} & 20.7\ci{2.3} & 1.8\ci{0.4} & 4.4\ci{0.5} \\
        DreamZero (DROID) & 500 & 0.7\ci{0.4} & 2.5\ci{0.9} & 1.6\ci{0.9} & 1.6\ci{0.8} & 13.4\ci{1.6} & 40.3\ci{2.8} & 5.5\ci{0.6} & 10.0\ci{0.7} \\
        MolmoAct2-DROID & 500 & 1.5\ci{0.5} & 2.9\ci{1.0} & 2.6\ci{1.2} & 2.0\ci{0.9} & 19.4\ci{1.9} & 60.0\ci{2.8} & 8.0\ci{0.8} & 14.7\ci{0.8} \\
        $\pi_{0.5}$-DROID & 500 & 1.9\ci{0.6} & 5.1\ci{1.3} & 6.3\ci{1.8} & 2.7\ci{1.0} & 23.6\ci{2.0} & 69.2\ci{2.6} & 10.1\ci{0.9} & 17.8\ci{0.9} \\
        UMI & 500 & 23.4\ci{1.8} & 7.6\ci{1.6} & 6.8\ci{1.8} & 6.7\ci{1.6} & 24.5\ci{2.1} & 45.2\ci{2.9} & 20.3\ci{1.1} & 21.1\ci{0.9} \\
        UMI-on-Air & 500 & 24.9\ci{1.9} & 7.6\ci{1.6} & 9.4\ci{2.1} & 6.3\ci{1.5} & 24.1\ci{2.1} & 44.3\ci{2.8} & 20.8\ci{1.1} & 21.4\ci{0.9} \\
        UMI (all views) & 500 & 65.4\ci{2.1} & 48.7\ci{3.0} & 39.8\ci{3.5} & 47.5\ci{3.1} & 60.6\ci{2.4} & 68.9\ci{2.7} & 60.1\ci{1.4} & 57.9\ci{1.1} \\
        \midrule
        GPT-6 Astra & 200 & \hphantom{ }6.9\ci{4.7} & \hphantom{ }0.0\ci{0.0} & -- & -- & 17.2\ci{7.0} & -- & \hphantom{ }9.6\ci{3.4} & \hphantom{ }9.6\ci{3.4} \\
        GPT-6 Astra & 500 & 44.8\ci{9.2} & 26.7\ci{11.4} & -- & -- & 44.8\ci{9.2} & -- & 41.1\ci{5.8} & 41.1\ci{5.8} \\
        GPT-6 Astra & 1000 & 48.3\ci{9.3} & 46.7\ci{12.9} & -- & -- & 65.5\ci{8.8} & -- & 54.8\ci{5.8} & 54.8\ci{5.8} \\
        \midrule
        \textbf{\modelshort} & 200 & 58.7\ci{2.1} & 22.0\ci{2.5} & 20.9\ci{2.9} & 22.7\ci{2.6} & 39.4\ci{2.4} & 56.7\ci{2.8} & 43.9\ci{1.4} & 40.9\ci{1.1} \\
        \textbf{\modelshort} & 500 & \textbf{70.4}\ci{2.0} & \textbf{59.6}\ci{2.9} & \textbf{48.7}\ci{3.6} & \textbf{54.5}\ci{3.1} & \textbf{60.9}\ci{2.3} & \textbf{71.8}\ci{2.6} & \textbf{64.7}\ci{1.4} & \textbf{63.0}\ci{1.1} \\
        \bottomrule
    \end{tabular}
\end{table*}

\begin{table}[t]
    \caption{\textbf{Where the real-world evaluation arms, OpenYAM and Franka, sit in the training corpus}. In addition, our OpenYAM arm carries a fine-tip flange gripper unseen during training. $P_k$ indicates \textit{percentile}.}
    \label{tab:yam_distribution}
    \centering
    {\footnotesize
    \setlength{\tabcolsep}{3.5pt}
    \begin{tabular}{lrrrrr}
        \toprule
                                 & \multicolumn{2}{c}{\textbf{OpenYAM}} & \multicolumn{2}{c}{\textbf{Franka}} & \\
        \cmidrule(lr){2-3} \cmidrule(lr){4-5}
                                 & \textbf{Value} & \textbf{$\mathbf{P_k}$} & \textbf{Value} & \textbf{$\mathbf{P_k}$} & \textbf{Median} \\
        \midrule
        Reach (m)                & 0.76   &  1.2 & 1.21    & 43.3 & 1.26   \\
        Arm mass (kg)            & 3.6    &  3.7 & 18.3    & 94.9 & 10.2   \\
        Joint stiffness          & 30     &  3.3 & 3500    & 90.0 & 1216   \\
        Damping gain             & 1.75   &  2.0 & 350     & 85.4 & 102    \\
        Armature                 & 0.017  & 15.6 & 0.100   & 44.8 & 0.130  \\
        Force limit              & 19     &  1.2 & 100,000 & 82.2 & 2293   \\
        \midrule
        Stiffness:inertia        & 3400   & 11.2 & 35,000  & 95.2 & 11,300 \\
        \midrule
        Dist.\ to nearest emb.\  & 0.545  & 98.9 & 0.065   & 15.8 & 0.116  \\
        \bottomrule
    \end{tabular}\par}
    \vspace{0.8em}
    % stacked in the same float so the figure stays next to its table
    \includegraphics[width=\linewidth]{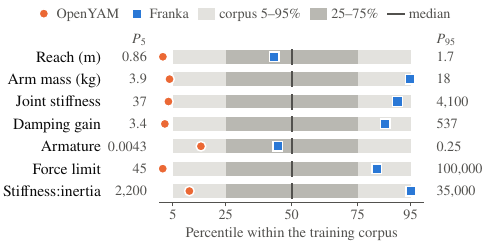}
    \captionof{figure}{\textbf{Percentile of each real-world arm within the training corpus} for each physical quantity in \Cref{tab:yam_distribution}. OpenYAM is at the edge of the distribution while Franka is a mix between in-distribution and the edge of the distribution.}
    \label{fig:arm_ranges}
\end{table}

\begin{table}
    \caption{\textbf{Real-world deployment on physical arms Franka and OpenYAM}. In the MolmoAct2 row, MolmoAct2's specialized models for each embodiment are used in order to measure its best performance. \textbf{N/S} indicates that a model fails or behaves completely unsafely on all attempts on an embodiment due to the lack of support for that embodiment.}
    \vspace{0.2em}
    \label{tab:real_world_results}
    \scriptsize
    \setlength{\tabcolsep}{2pt}
    \centering
    \begin{tabular}{l|ccc|cc}
        \toprule
        & \multicolumn{3}{c|}{\textbf{Success}} & & \raisebox{-1ex}[0pt][0pt]{\textbf{Median}} \\
        \cmidrule(lr){2-4}
        \textbf{Model} & \textbf{Franka} & \textbf{OpenYAM} & \textbf{Overall (\%)} & \textbf{Params} & \textbf{steps} \\
        \midrule
        \multicolumn{6}{c}{\textit{Some or all target embodiments seen}} \\
        \midrule
        MolmoBot                  & \hphantom{0}8/20 & \textit{N/S} & \hphantom{0}8/40 (20\%)           & 4B   & 142 \\
        LAP-3B                    & \hphantom{0}8/20 & 0/20         & \hphantom{0}8/40 (20\%)           & 3B   & 241 \\
        $\pi_{0.5}$               & 13/20            & \textit{N/S} & 13/40 (33\%)                      & 3B   & 165 \\
        DreamZero                 & \textbf{15/20}   & \textit{N/S} & 15/40 (38\%)                      & 14B  & 194 \\
        MolmoAct2                 & 14/20            & 7/20 & 21/40 (53\%)                      & 5B   & 356 \\
        \midrule
        \multicolumn{6}{c}{\textit{No target embodiments seen}} \\
        \midrule
        UMI                       & \hphantom{0}1/20 & \hphantom{0}0/20 & \hphantom{0}1/40 \hphantom{0}(3\%) & 0.6B & 152 \\
        UMI-on-Air                & \hphantom{0}1/20 & \hphantom{0}0/20 & \hphantom{0}1/40 \hphantom{0}(3\%) & 0.6B & 298 \\
        UMI (all views)           & 11/20            & \hphantom{0}0/20 & 11/40 (28\%)                      & 0.6B & 166 \\
        \midrule
        \textbf{\modelshort}      & 14/20            & \textbf{13/20}   & \textbf{27/40 (68\%)}             & 0.4B & 177 \\
        \bottomrule
    \end{tabular}
\end{table}

The key takeaways are as follows: (a) In simulation on robots seen to baselines, \model performs better or on par with existing methods despite not seeing the robot itself; (b) In simulation on unseen robots, \model outperforms all baseline methods, including embodiment-specific and embodiment-agnostic policies, as well as the generalist VLM GPT-6 Astra, despite a smaller step budget; (c) In real-world deployment, \model performs better or on par with existing methods on robots these existing methods specialize towards; (d) \model adapts its motion strategy based on the embodiment and the task at hand, instead of retargeting one canonical grasp across bodies.

\vspace{0.3em}\noindent\textbf{Training data.} We train on our corpus of 1M procedurally generated manipulators (\Cref{sec:method_trajgen}) performing manipulation in MolmoSpaces scenes~\citep{kim2026molmospaceslargescaleopenecosystem}.

\vspace{0.3em}\noindent\textbf{Baselines.} We compare with frontier policies including $\pi_{0.5}$~\citep{intelligence2025pi05visionlanguageactionmodelopenworld}, DreamZero~\citep{ye2026worldactionmodelszeroshot}, MolmoAct2~\citep{fang2026molmoact2actionreasoningmodels}, GPT-6 Astra (max thinking), and embodiment generalization specialists UMI~\citep{chi2024universalmanipulationinterfaceinthewild} and UMI-on-Air~\citep{gupta2026umionairembodimentawareguidanceembodimentagnostic}. To address the possibility that a multi-view setup would provide better visual context compared to UMI's single-view setup, we add a variant of UMI supplied with a 3-view camera setup as another baseline. We train UMI on the same simulated trajectories as our training but with UMI's setup, to measure the effects of our architectural choices that facilitate our specific input conditioning (\Cref{sec:method_embodiment}). In real-world deployment, we additionally measure LAP-3B~\citep{zha2026laplanguageactionpretrainingenables}, a zero-shot cross-embodiment transfer model, at their released checkpoint.

\vspace{0.3em}\noindent\textbf{Benchmarks.} We evaluate policies in three settings: i) on the existing MolmoSpaces benchmark that uses a robot that is in-distribution for the baseline policies, ii) on \modelbench, our benchmark based on held-out training data so unseen to all policies, and iii) on real-world tasks.

\vspace{0.3em}\noindent\textbf{Implementation details.} \model is a 400M-parameter flow-matching policy over a fixed 12-slot node and action layout. It observes RGB views encoded by a frozen SigLIP~2 image encoder~\citep{tschannen2025siglip2multilingualvisionlanguage}, language encoded by the frozen text backbone of Qwen2.5-VL~\citep{bai2025qwen25vltechnicalreport}, and the per-node embodiment representation of \Cref{sec:method_embodiment}. Training samples are 32-frame windows with a 31-step action prediction horizon. We train with an effective batch size of 128 with a learning rate of $10^{-3}$ for 520K steps. At evaluation the policy predicts a 31-step chunk and executes the first 8 steps every chunk.

\subsection{Evaluation on seen robots in simulation}
\label{sec:franka_sim}

We evaluate \model in simulation on the MolmoSpaces benchmark, which uses the Franka Panda in the DROID configuration. All baselines have seen the Franka robot during training, whereas \model does not.

\vspace{0.3em}\noindent\textbf{\model performs on par with embodiment-specialized models on their expert embodiments.} \Cref{tab:franka_sim}(full scenes) reports success on $2{,}622$ held-out episodes under the same $500$-step cap. \modelshort reaches $69.1\%$ overall, against $63.0\%$ for MolmoBot and under $30\%$ for the DROID-trained policies, and leads on four of the six task families, with the largest margins where the task constrains placement or approach: for example, $+15.5$ percentage points over MolmoBot on placing next to a reference object.

\vspace{0.3em}\noindent\textbf{\model is robust against changes in object, lighting, and prompt format, and handles changes in camera and robot pose about as well as baselines.} $\pi_{0.5}$ and MolmoAct2 both score below their MolmoSpaces' reported success. To understand this difference, we notice that MolmoSpaces' scenes, especially the ones involving locating an object, have increased in difficulty since release~\cite{deshpande2026molmob0tlargescalesimulationenables}. We thus label our benchmark as \Cref{tab:franka_sim}(full scenes) and contrast with \Cref{tab:franka_sim}(simple scenes). Other than adding harder pick objects and tasks, the change also includes more diverse and randomized lighting, camera, robot pose, and prompt. To find out why $\pi_{0.5}$ and MolmoAct2 drop in performance, we gradually change each of the variables in the Pick task and contrast the results with \model in \Cref{fig:ms_pick_cumulative}, which shows that \model fluctuates within error margins while $\pi_{0.5}$ and MolmoAct2 drop.

\subsection{Evaluation on unseen robots in simulation}

Next we want to test how well baseline methods generalize to novel embodiments. We use the held-out \modelbench benchmark (\Cref{sec:method_scaling_embodiments}).

\vspace{0.3em}\noindent\textbf{\model generalizes to unseen embodiments in simulation.} Rollout success on a random subset of 2000 episodes sampled from our benchmark is reported in \Cref{tab:molmospaces_benchmark_results}. From the results, we can see that \model outperforms baselines at the same number of rollout steps, leading on six task families, and its overall success rate of 63.0\% is 5.1 percentage points above the next best method, UMI with all camera views. Beyond policies, the general-purpose GPT-6 Astra model also struggles with random embodiments, trailing \model by 23.6 percentage points at the same step count on the task families both were evaluated on (41.1\% vs.\ 64.7\%). We empirically observe that GPT-6 Astra, thanks to its language and multimodal reasoning capabilities, persistently retries with different solutions; however, even at double the step count, it is still behind \model by 9.9 percentage points.

\subsection{Evaluation on real robots}
\label{sec:real_world}

We deploy \model on two physical arms without fine-tuning: a Franka Panda, the robot every baseline was trained on, and an OpenYAM arm, which only MolmoAct2-BimanualYAM has trained on and which carries a fine-tip flange gripper that no policy has seen.

\vspace{0.3em}\noindent\textbf{Franka.}
\label{sec:franka_real}
We deploy on Franka Panda in the DROID configuration, the robot on which every baseline is also trained. For the MolmoAct2 baseline, its DROID-specific variant is deployed to ensure maximum compatibility.

\vspace{0.3em}\noindent\textbf{OpenYAM.}
We next deploy on a physical OpenYAM arm fitted with its fine-tip flange gripper. For the MolmoAct2 baseline, the BimanualYAM variant is deployed and both YAM arms are enabled with their standard camera setup to comply with its model's expectation.

OpenYAM is in a relatively low-density region of our training data, and its fine-tip flange gripper is not seen during training. \Cref{tab:yam_distribution} and \Cref{fig:arm_ranges} show that many of its key characteristics lie beyond the 5th/95th percentile.

\vspace{0.3em}\noindent\textbf{On real unseen robots, \model outperforms other models not specialized to the embodiment, and is on par with embodient-specialized models.}
We measure the success rate on an object pick-and-place task, varying the location of the object and the target within and at the boundary of the embodiment's reach. We measure 20 starting setups per robot; in each setup, every policy starts from the same object positions, under a $1000$ step horizon (\Cref{tab:real_world_results}). \modelshort is deployed zero-shot; it receives no fine-tuning on either robot, real or simulated.

On Franka, \modelshort succeeds on $14$ of $20$ setups, the same as MolmoAct2-DROID and on par with the policies trained on real DROID data from this robot. On OpenYAM, it succeeds on $13$ of $20$ setups, while MolmoAct2, which is specifically trained on YAM, succeeds on only $7$. Empirically, MolmoAct2-BimanualYAM fails frequently when the object is farther from the robot compared to the receptacle, or partially hidden behind the receptacle. Despite LAP-3B's 30\% success rate on I2RT YAM, it does not succeed on OpenYAM and struggles at the pick phase.

Notably, while UMI with all views achieves $11/20$ on Franka, it fails on all OpenYAM setups, which fall on the edge of the training distribution. UMI (all views) receives the same training data as our model except for the scene-grounded kinematic graph (\Cref{sec:method_embodiment}).

\subsection{Motion strategy analysis}
\label{sec:motion_strategy}

During evaluation, we notice that when the object pickup site is close to the receptacle, \model sometimes takes advantage of OpenYAM's long link leading up to the gripper, simply rotating the gripper in the Z axis to reach the receptacle instead of lifting and dropping.

This is plausibly a side-effect of our data generation method: we allow for poses at different phases to have different orientations; thus, when the receptacle is close by, rotating is a viable, if not more direct, strategy. Because every trajectory in our corpus is one that its own embodiment could execute, the
data does not commit to a single grasp retargeted to different embodiments: depending on the body and the object, successful grasps approach from above, from the side, or in between. We examine whether
\model carries the flexibility to change its strategy depending on the embodiment and the task.

\vspace{0.3em}\noindent\textbf{Setup.} We perform this experiment on OpenYAM, a normal-sized arm, and Lite6, an arm relatively limited in reach. A knife, a cube, and a bowl are visible. The knife lies on the surface and cannot be picked up easily from the top.

\begin{figure}[t]
    \centering
    \begin{minipage}[t]{0.495\linewidth}
        \centering
        \includegraphics[width=\linewidth]{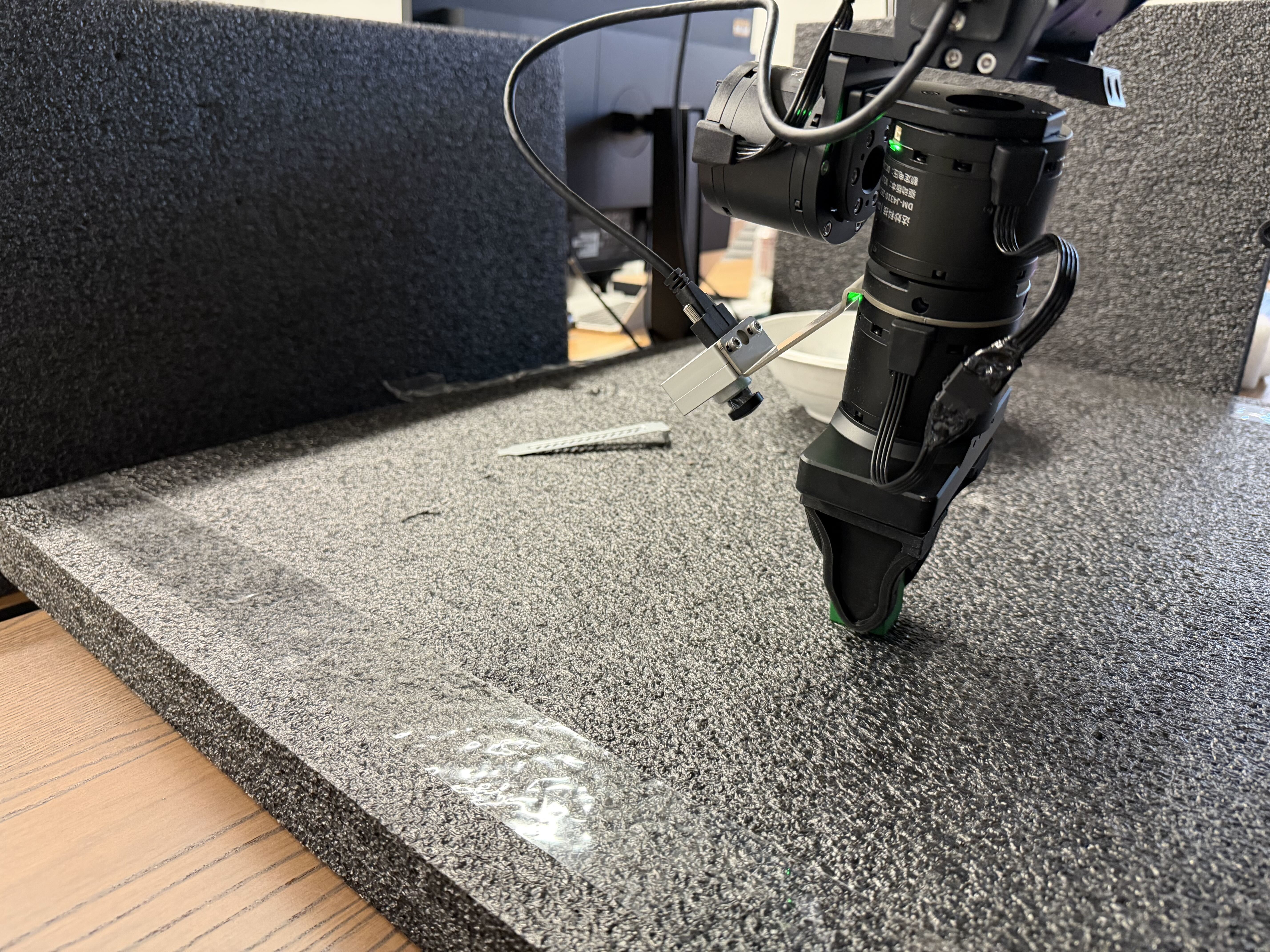}\\[1pt]
        \includegraphics[width=\linewidth]{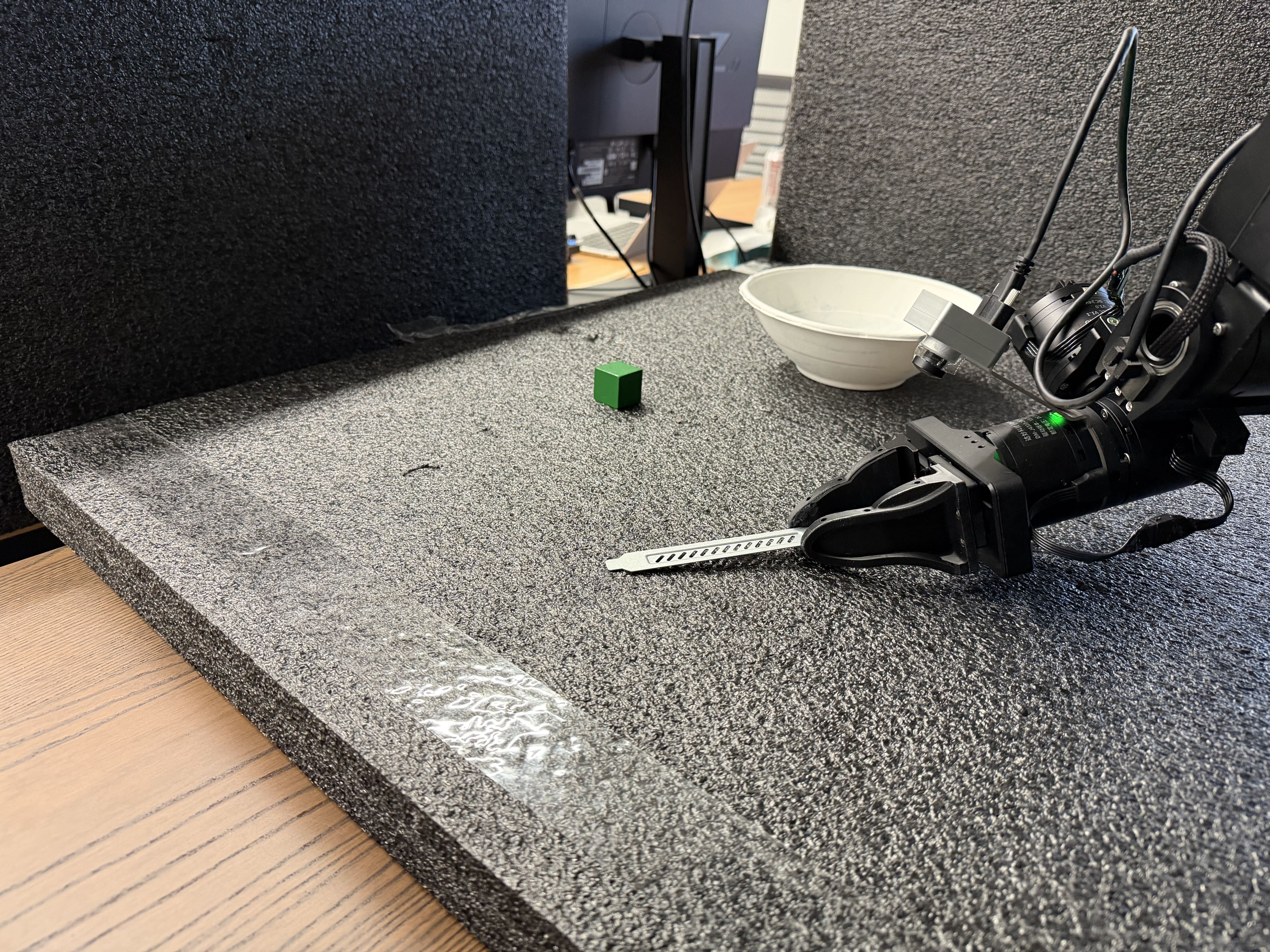}\\[1pt]
        {\small (a) Same emb., different tasks.}
    \end{minipage}\hfill
    \begin{minipage}[t]{0.495\linewidth}
        \centering
        \includegraphics[width=\linewidth]{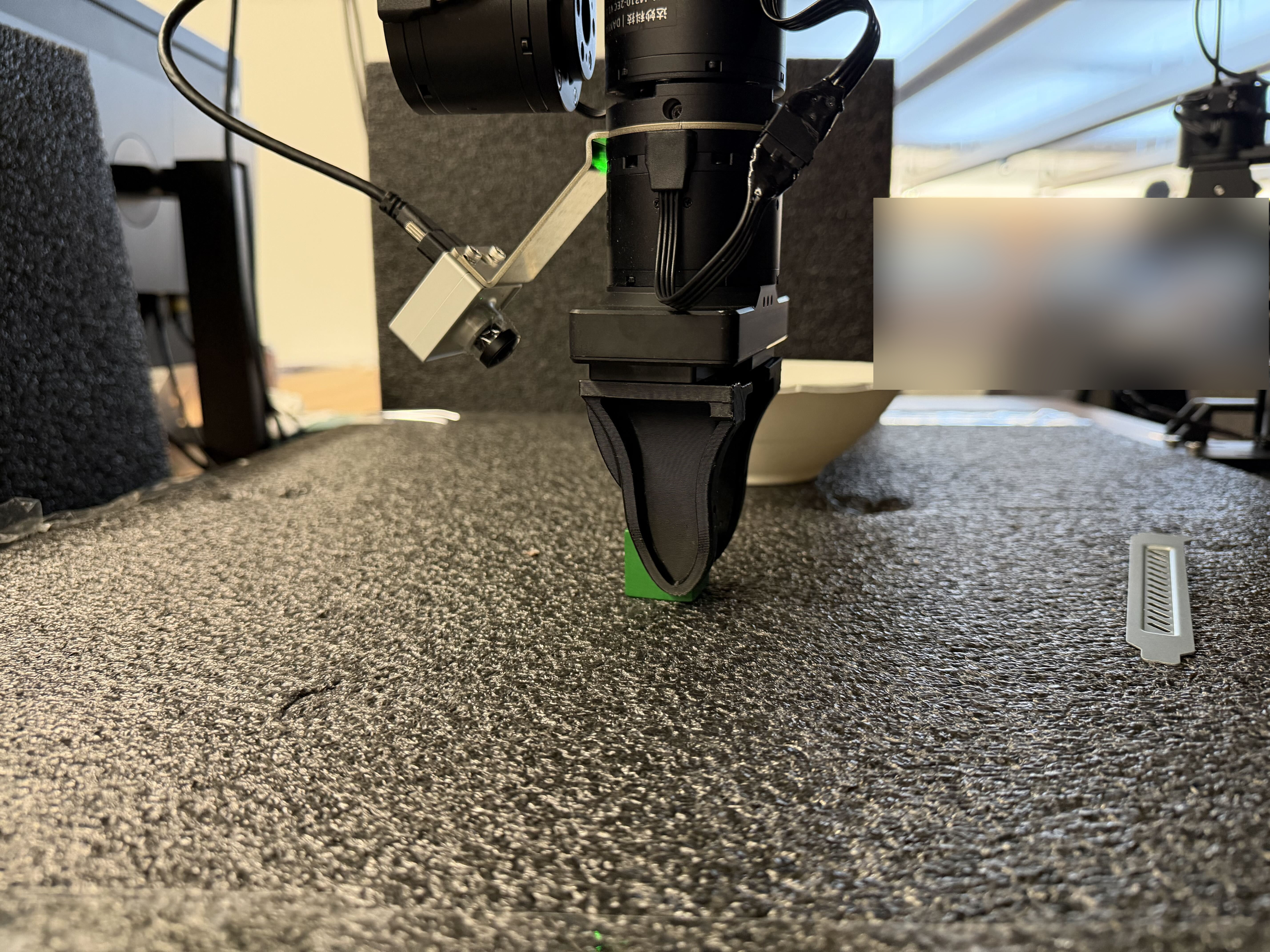}\\[1pt]
        \includegraphics[width=\linewidth]{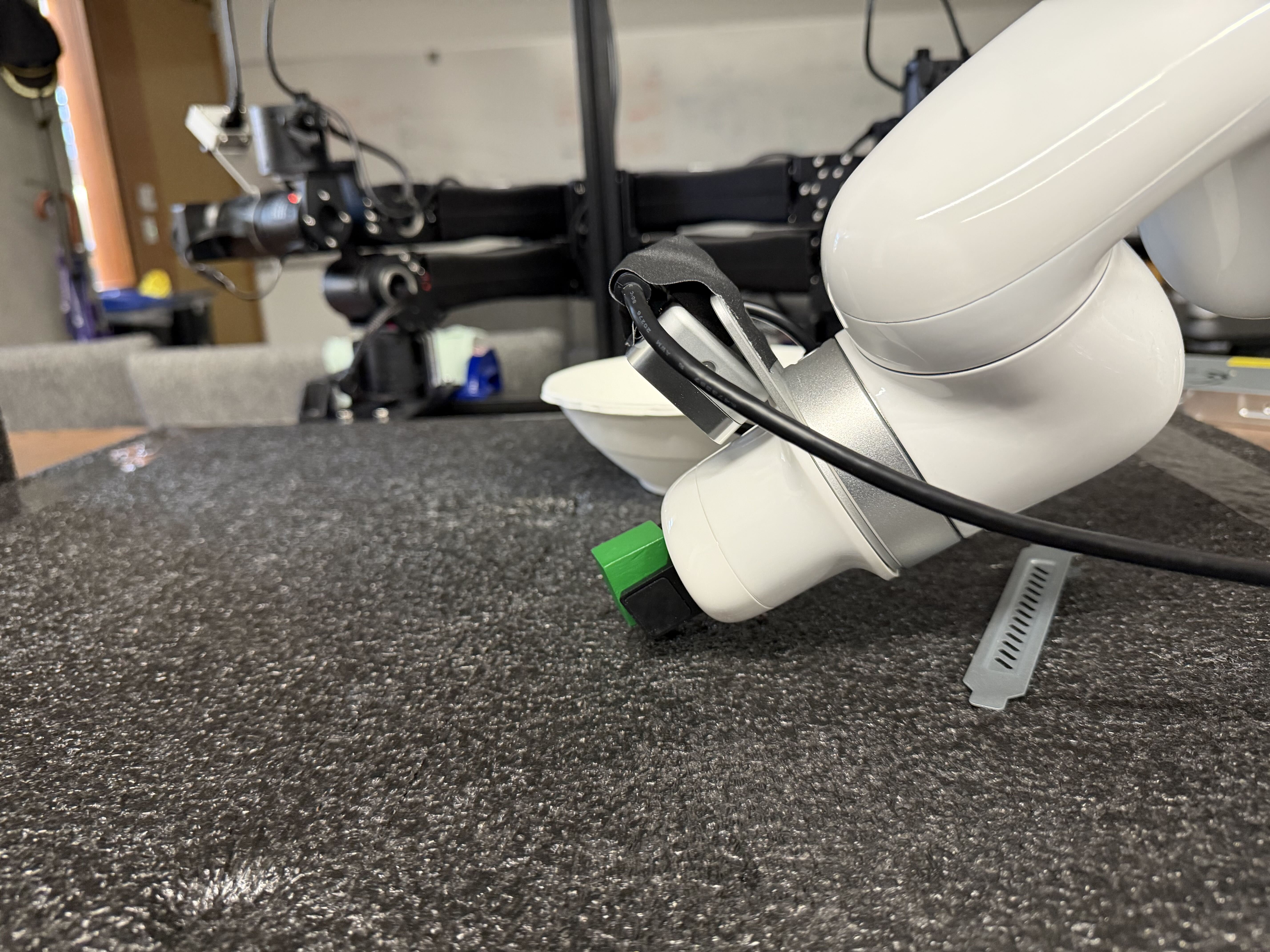}\\[1pt]
        {\small (b) Same task, different embs.}
    \end{minipage}
    \caption{\textbf{\model chooses its grasp orientation for the task and the embodiment.} (a) Same embodiment, different tasks: top-down for the cube, side grasp for the knife. (b) Same task, different embodiments: top-down on OpenYAM, lateral on Lite6.}
    \label{fig:motion_strategy}
\end{figure}

\vspace{0.3em}\noindent\textbf{On real tasks, \model adapts its motion strategy to the task.}
Figure~\ref{fig:motion_strategy}(a) shows two grasps by the same OpenYAM arm from the same deployment. For the cube (top), \model approaches from above and closes the gripper vertically. For the knife (bottom), it instead pitches the wrist until the gripper is level with the table and closes on the blade from the side, the orientation in which the thin body fits between the fingertips with maximum contact area. The policy treats the wrist as a free degree of freedom and picks the grasp orientation the object calls for, rather than holding a fixed top-down pose.

\vspace{0.3em}\noindent\textbf{On real robots, \model adapts its motion strategy to the embodiment.}
Figure~\ref{fig:motion_strategy}(b) fixes the task and varies the body. When the object is placed close to the robot, both OpenYAM and Lite6 approaches normally from the top. However, when the object is placed at the edge of Lite6's relatively short reach (bottom), it approaches from the side with the gripper lying about $20^\circ$ from the table, while on OpenYAM (top) it approaches from the top as OpenYAM can easily reach the same distance. Additionally, when attempting to pick from the side, the angles of approach are slightly different between OpenYAM and Lite6 (\Cref{fig:motion_strategy}(a)(b) bottom), such that both robots would be lying as flat as possible without putting excessive force on the table, which would trigger a gripper collision in simulation and a collision stop in real drivers.

\vspace{0.3em}\noindent Overall, we can see that \model adapts its motion strategy to the embodiment and the task at hand, rather than retargeting one canonical grasp across bodies and forced into an infeasible pose on one of them.

\subsection{Ablation study}

\begin{figure}[t]
    \centering
    \includegraphics[width=\linewidth,height=2.25in,keepaspectratio]{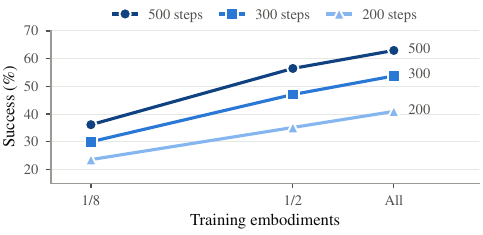}
    \caption{\textbf{Success vs.\ number of training embodiments.} Success on the same held-out episodes as \Cref{tab:molmospaces_ablation}, at each rollout step cap. Standard errors ($\leq 1.1$ points) are smaller than the markers.}
    \label{fig:embodiment_scaling}
\end{figure}

\paragraph{Scaling training embodiments helps.}
We subsample the generated training embodiments to one half and one eighth of the corpus (400K and 100K, respectively), holding architecture and training schedule fixed (\Cref{tab:molmospaces_ablation} \& \Cref{fig:embodiment_scaling}). Success falls monotonically: $-6.5\%$ at one half, and $-26.8\%$ at one eighth, on a 500 rollout step limit.

\vspace{0.3em}\noindent\textbf{Scene-grounded kinematic graph and camera-view inputs are important to success.}
We then remove inputs at fixed data (\Cref{tab:molmospaces_ablation}). Dropping the per-node kinematic graph tokens costs $5.1$ percentage success: the body is partially inferable from pixels and proprioception, but not fully, demonstrating the advantage of supplying the scene-grounded graph through camera-grounded nodes instead of through robot-frame proprioception. Removing the exocentric views as well, leaving only wrist observations, costs a further $36.8\%$.

\begin{table}[t]
    \caption{\textbf{Ablation study.} Success rates on the same held-out episodes and hyperparameters.}
    \label{tab:molmospaces_ablation}
    \small
    \setlength{\tabcolsep}{3pt}
    \centering
    \begin{tabularx}{\columnwidth}{>{\raggedright\arraybackslash}X c c c}
        \toprule
        & \multicolumn{3}{c}{\textbf{Success \% with rollout steps ...}} \\
        \cmidrule(lr){2-4}
        \textbf{Model} & \textbf{200} & \textbf{300} & \textbf{500} \\
        \midrule
        \textbf{\modelshort}                     & \hphantom{$-$}40.9\ci{1.1} & \hphantom{$-$}53.8\ci{1.1} & \hphantom{$-$}63.0\ci{1.1} \\
        \midrule
        w/ 1/2 embodiments                  & \hphantom{0}$-$5.8\ci{1.1} & \hphantom{0}$-$6.7\ci{1.2} & \hphantom{0}$-$6.5\ci{1.2} \\
        w/ 1/8 embodiments                  & $-$17.4\ci{1.2} & $-$23.8\ci{1.2} & $-$26.8\ci{1.3} \\
        w/o scene grnd.                  & \hphantom{0}$-$4.6\ci{1.1} & \hphantom{0}$-$6.0\ci{1.2} & \hphantom{0}$-$5.1\ci{1.2} \\
        \makecell[tl]{w/o scene grnd.\ \& \\ w/o exo views} & $-$29.0\ci{1.1} & $-$37.9\ci{1.2} & $-$41.9\ci{1.3} \\
        \bottomrule
    \end{tabularx}
    \vspace{-1em}
\end{table}

\section{Discussion}
We find that scaling diverse robotic embodiments and supplying scene-grounded kinematic graphs lead to better generalization to unseen robots on real-world tasks, enabling body-specific motion strategies for the task at hand. Scaling the number of training embodiments drives performance on held-out simulated embodiments. The scene-grounded kinematic graph provides a small amount of improvement as well, but farther away from the training distribution, the graph's contribution appears much larger (\eg OpenYAM). These results suggest that embodiment scale covers bodies that resemble the training ones, while the graph carries the policy to a body at the edge of that distribution.

An open question is to scale embodiments types further, including bimanual arms, humanoids, and everyday objects such as tools, equipment, and vehicles.

\section*{Acknowledgments}
We thank Jiawei Gu for the support of figure drafts and providing feedback, and Lindsey Li for helping proofread the work. This work used resources through allocation \#CIS261283 from the Advanced Cyberinfrastructure Coordination Ecosystem: Services \& Support (ACCESS) program, which is supported by U.S.\ National Science Foundation grants \#2138259, \#2138286, \#2138307, \#2137603, and \#2138296. This work was supported in part by Advanced Micro Devices, Inc.\ under the AMD University Program's AI \& HPC Cluster. This work was supported in part by the NSF through award IIS-2211133.

{
    \small
    \bibliographystyle{ieeenat_fullname}
    \bibliography{main}
}


\end{document}